## METHODS

# Methods and Strategies for Improving the Novel View Synthesis Quality of Neural Radiation Field

SHUN FANG[1], (Senior Member, IEEE), MING CUI[2], XING FENG[2], AND YANNA LV[1]
[1]Lumverse Research Institute, Beijing Lumverse Technology Company Ltd., Beijing 100043, China
[2]Beijing Lumverse Technology Company Ltd., Beijing 100043, China

Corresponding author: Ming Cui (cuiming@lumverse.com)

This work was supported by Beijing Municipal Science and Technology Commission and Zhongguancun Science and Technology Park Management Committee of Commerce under Grant Z221100007722005 and Grant 20230468276.

**ABSTRACT** Neural Radiation Field (NeRF) technology can learn a 3D implicit model of a scene from 2D images and synthesize realistic novel view images. This technology has received widespread attention from the industry and has good application prospects. In response to the problem that the rendering quality of NeRF images needs to be improved, many researchers have proposed various methods to improve the rendering quality in the past three years. The latest relevant papers are classified and reviewed, the technical principles behind quality improvement are analyzed, and the future evolution direction of quality improvement methods is discussed. This study can help researchers quickly understand the current state and evolutionary context of technology in this field, which is helpful in inspiring the development of more efficient algorithms and promoting the application of NeRF technology in related fields.

**INDEX TERMS** 3D reconstruction, new view synthesis, neural radiation field, volume rendering.

## I. INTRODUCTION

New view synthesis and 3D modeling are key research areas in computer vision. In recent years, researchers have increasingly focused on the NeRF [1] in this regard. NeRF provides an accurate and simple method to represent 3D scenes useing an implicit function based on MLPand, and has achieved satisfactory rendering quality in 3D reconstruction tasks. Current efforts aim to extend the original NeRF to different situations, such as scene synthesis [2], [3], dynamic scenes [4], [5], large scene reconstruction [6], [7] or rapid convergence [8], [9], among others.

Since NeRF was published in 2020, the NeRF paper has been cited more than thousands of times in the past three years. In addition, researchers have made numerous improvements to the NeRF technology. Some work have focused on optimizing the rendering speed of NeRF [10], [11], while others have explored different application scenarios

The associate editor coordinating the review of this manuscript and approving it for publication was Rosalia Maglietta.

[12], [13]. Futhermore, there have been efforts to extended NeRF for scene inpainting [14], [15], texture synthesis [16], handing complex scenes [17], and addressing more challenging problems.

This article reviews several situations where the quality of NeRF image rendering needs to be improved: For example, training on sparse input views can lead to overfitting and incorrect scene depth estimation, resulting in artifacts in the rendered new views. NeRF can cause blurry or distorted renderings when reflective and refractive objects are present. When the input image scene becomes complex, the MLP structure using the basic original NeRF is not enough to render high-quality new perspective images, and the 3D scene cannot maintain the appearance consistency and geometric rationality of multi-views, resulting in low quality synthetic images. It is difficult to directly model subtle mesoscopic structures because NeRF's inherent multi-layer perceptron has difficulty learning high-frequency details, and there are blur problems in high-resolution rendering. Problems with rendering quality hinder the deployment of NeRF technology in practical applications. The specific reasons for the blurred image rendering quality of NeRF are summarized as follows:

### A. INSUFFICIENT TRAINING DATA
The NeRF model requires a large amount of high-quality training data to learn the lighting and depth information of the scene. If the training data is limited or of poor quality, the model may not capture rich scene details, resulting in lower image quality.

### B. SCENE COMPLEXITY
The basic NeRF can only handle static images. When dealing with dynamic scenes, large scenes, unbounded scenes, or complex scenes, the synthesized images will have artifacts such as ghosting, blurring, and inconsistent views.

### C. IMAGE RESOLUTION
If the resolution of the input image is low, it can affect the quality of the final reconstruction and synthesized images from new viewpoints. However, inputting high-resolution images into the basic NeRF can lead to computational explosions.

Based on the current problems of NeRF in image rendering quality, this article provides a review of NeRF's latest research technologies to improve rendering quality. It not only introduces variety a lot of methods to improve rendering quality, but also analyzes and summarizes the characteristics and limitations of these methods. Then we analyze the possibility of combining distinct technologies to achieve improved quality effects, which will help stimulate the generation of more efficient algorithms. The ultimate goal is to advance the application of NeRF technology in new view synthesis and other fields. Figure 1 summarizes the four core aspects of improving image rendering quality introduced in this article.

## II. PRINCIPLE OF NEURAL RADIATION FIELD
For a given three-dimensional scene, the appearance of any location depends on the specific location and viewing angle. The colors displayed by the scene are related to lighting conditions, causing the colors to change when the same location is viewed from different angles. NeRF represents a three-dimensional scene as a radiation field approximated by a neural network. The radiance field describes the volumetric density and color of each point in the scene for every viewing direction, and represents the static scene as a continuous five-dimensional vector function, expressed as follows.

$$(r, g, b, \sigma) = F_\Theta(x, y, z, \theta, \varphi) \quad (1)$$

$(x, y, z)$ represents the three-dimensional coordinates in the scene space, $d = (\theta, \phi)$ represents the viewing direction of the input image, $c = (r, g, b)$ represents the color emitted from this position in the $d$ direction. $\theta$ represents volume density, which is similar to differentiable opacity and represents the amount of radiation accumulated when light passes through $(x, y, z)$. $F_\Theta$ means that the five-dimensional function is implicitly expressed using one or more MLP neural networks. By inputting a series of images taken from different perspectives, the output is the color $c$ and volume density $\sigma$ corresponding to the three-dimensional space position. Based on the scene representation foundation of NeRF, classical volume rendering methods can be employed to render new images from different perspectives. Specifically, for any pixel in the image, $N$ points are sampled along the ray $r$ of the observation angle. For each sampling point, $\Theta_i$ and $rgb_i$ are first calculated based on $F_\Theta$, and then calculated according to the following formula Final color value:

$$\hat{rgb} = \sum_{i=1}^{N} rgb_i w_i \quad (2)$$

$$w_i = T_i \alpha_i \quad (3)$$

$$T_i = \exp\left(-\sum_{j=1}^{i-1} \sigma_j \delta_j\right) \quad (4)$$

$$\alpha_i = 1 - \exp(-\sigma_i \delta_i) \quad (5)$$

$\delta_i$ represents the sampling interval on ray $r$.

In order to train the specific parameters of the MLP corresponding to $F_\Theta$, for a given scene, $n$ images are captured using cameras with different postures. By using the gradient descent method, fitting $F_\Theta$ by minimizing the error between the predicted image $I_p$ and the ground truth image $I_c$, that is $\min \sum_{i=1}^{m} |I_p - I_c|^2$.

Figure 2 shows the flow of the NeRF algorithm. In the literature [1], in order to obtain an output image containing more high-frequency information, the author performed a high-order encoding operation $\gamma(\cdot)$ on the position and perspective parameters of the input MLP.

## III. AN INTRODUCTION TO TECHNIQUES TO IMPROVE RENDERING QUALITY
In view of the huge potential development prospects of NeRF, researchers have proposed a series of technologies to improve the quality of rendered images in the past three years, so that NeRF can be combined with more algorithms and models to handle different tasks. As analyzed in the previous section, the actual rendering quality of NeRF is comprehensively affected by factors such as the amount of training data, the resolution of the input image, and the complexity of the application scenario. This article categorizes and introduces the latest techniques from four aspects: prior guidance, adjusting NeRF model structure, adopting pre-trained models, and high-resolution image processing.

### A. PRIOR GUIDANCE
In 3D surface reconstruction and new view synthesis, prior guidance refers to the process of assisting relative tasks through known prior knowledge or models to improve the accuracy and stability of task results. The prior information can be object shape, texture, posture, motion, etc. Prior guidance plays an important role in 3D surface reconstruction and new view synthesis, improving the quality and accuracy of the results and shortening the required time and computational resources. Different prior guidance can be selected and combined according to specific application scenarios and requirements. In a series of NeRF papers, many researchers have utilized prior knowledge to achieve rendering in different application scenarios, ultimately achieving higher visual quality.

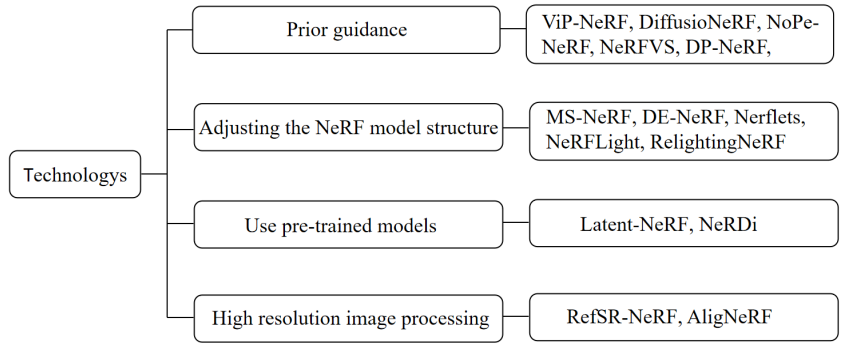

**FIGURE 1.** Technical application to improve NeRF image rendering quality.

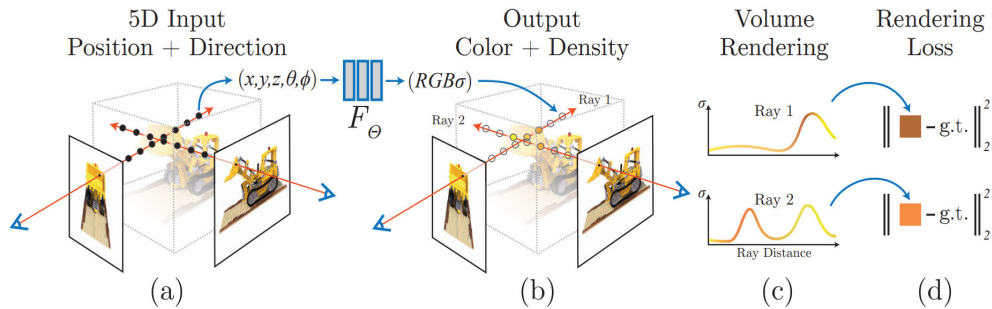

**FIGURE 2.** Neural Radiation Field Algorithm Process [1].

It is known that each scene of NeRF requires at least 50-100 view inputs for training in order to synthesize new views from arbitrary viewpoints. However, when the number of input views is not allowed, training will lead to artifacts in the rendered new views. Therefore, introducing prior guidance can solve such problems and improve the rendering quality of the images. ViP-NeRF [18] and DiffusioNeRF [19] both introduce prior knowledge and combine it with NeRF to improve the rendering quality of synthesized images from new viewpoints under sparse input view conditions.

ViP-NeRF computes the visibility priors between views by assuming that the visibility of pixels in different input views can provide a more reliable estimate to supervise the density. The specific implementation steps are as follows: 1) visibility regularization; 2) visibility prior; 3) effective prediction of visibility. Given a main view and any secondary view, the visibility prior is used to estimate whether each pixel in the main view is also visible in the secondary view. Visibility is computed using planar scanned volumes, which does not require any pre-training. By regularizing NeRF with visibility priors, NeRF is successfully trained with few input views, as shown in Figure 3. The plane scan volume is created by warping the different depth planes of the secondary view image to the main view, and compared with the main view image to obtain the error map. It can be found from the figure that the second and third planes match the main view better than the other two planes, resulting in lower errors. The minimum error of all planes is then thresholded to obtain a visibility prior map corresponding to the primary view image. Finally, on the RealEstate-10K data set, the PSNR index of ViP-NeRF is 3-4 points higher than the latest paper in 2, 3, and 4 input views.

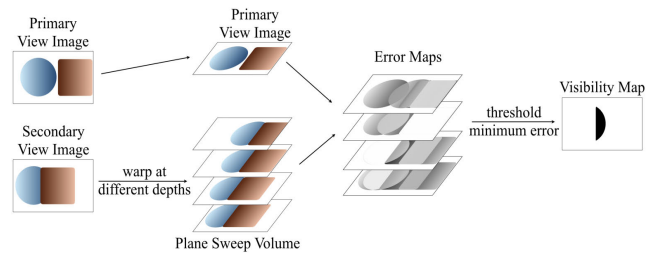

**FIGURE 3.** Visibility prior calculation [18].

DiffusioNeRF utilizes a denoising diffusion model to learn priors for scene geometry and color. A denoising diffusion model is trained on RGBD patches from the synthetic Hypersim dataset can be used to predict the logarithmic gradient of the joint probability distribution of color and depth patches. The logarithmic gradient of the priors for RGBD patches helps normalize the geometry and color of the scene. During NeRF training, random RGBD patches are rendered and the estimated gradients of the log-likelihood is backpropagated to the color and density fields. The learned prior achieves better quality in the reconstructed geometry. The PSNR indicators in the 3, 6, and 9 input views of the LLFF dataset are 1-2 points higher. Among the three input views, the PSNR index is nearly 1 point higher than ViP-NeRF.

**TABLE 1.** Priori guidance related papers three index values.

| Model | PSNR | SSIM | LPIPS |
|---|---|---|---|
| ViP-NeRF | 28.13 | 0.8588 | 0.1386 |
| DiffusioNeRF | 25.18 | 0.883 | 0.046 |
| NoPe-NeRF | 31.86 | 0.83 | 0.38 |
| NeRFVS | 26.946 | 0.891 | 0.268 |
| DP-NeRF | 25.70 | 0.7922 | 0.1405 |

NoPe-NeRF [20] introduces undistorted monocular depth prior to address the challenging problem of training a NeRF without precomputing camera poses. The prior is generated by correcting the shift parameters and scale during training and can be used to constrain the relative poses between consecutive frames, which is achieved using a novel loss function proposed in the paper. Temples and ScanNet outperform existing methods in terms of novel view rendering quality and pose estimation accuracy.

In terms of prior guidance, NeRFVS [21] proposes an overall prior for neural reconstruction to improve NeRF and realize free view synthesis in indoor scenes. DP-NeRF introduces physical scene priors to handle challenging scenes with blur and exposure [22], due to the sparsity of point cloud representation, utilizing point cloud 3D priors to synthesize high-quality images from colored point clouds is often used for novel indoor scenes [23]. These papers change the ideas of basic NeRF in different aspects, but overall improve the quality of NeRF rendered images. Table 1 shows the indicator data of the above papers.

### B. ADJUSTING THE NERF MODEL STRUCTURE

NeRF was initially applied to new viewpoint synthesis of static images, but as scenes become more complex, NeRF may need to adjust the basic MLP model structure to flexibly handle the task, such as changing the levels and number of MLPs to encode more high-frequency information, or modifying MLP input and output information to improve rendering quality.

MS-NeRF [24] addresses the rendering problem of distortion and blurring caused by reflection or refraction in NeRF. While NeRF satisfies the principle of multi-view consistency for new viewpoint synthesis, there is no consistency in the presence of mirrors in the scene. The paper proposes a multi-space neural radiation field. In order to understand the existence of refractive and reflective objects, the scene is represented by a set of characteristic fields in parallel subspaces. The specific model architecture is shown in Figure 4. where the output and volumetric rendering module of NeRF are modified to output density and features instead of density and color. Multiple sets of density and features are outputted. Volumetric rendering then generates multiple feature maps. Decoder MLP is applied in each subspace to decode RGB images, and Gate MLP obtains pixel-level weights for the feature maps. The features from each subspace are integrated by summation to obtain the final result. The advantage of MS-NeRF is that it can automatically handle mirror objects in 360Â° scene rendering. On synthetic data sets, real data sets and RFFR data sets, the PSNR index is 2-4 points higher.

In order to separate the appearance materials and lighting to better relight the high-frequency environment and allow the appearance to be edited, DE-NeRF [25] is proposed to decompose the lighting, geometry and appearance of the input scene. The network architecture is shown in Figure 4, which is divided into three modules: geometric reconstruction, scene decomposition, and relighting. The geometric reconstruction is to train a signed distance field (SDF) function by inputting a set of 2D images to reconstruct the display grid as the input scene of MLP. Different from NeRF, DE-NeRF decomposes the scene into geometric features and appearance features (including diffuse reflection, roughness, specular tint), combined with the distance $h$ from the sampling point to the grid, is input into four corresponding MLPs to predict diffuse albedo, roughness value, and specular tint. Then integrate the diffuse reflectance with the environment map to get the diffuse color $c_d$, train a specular lighting MLP $F_s$ to get the specular lighting $c_l$, and then multiply it with the specular hue to get the specular color $c_s$, $c_s + c_d$ gets the final color $c$. The final quality on the NeRF Synthetic and Shiny Blender data sets is 2 points higher than the PSNR index compared to the latest papers.

NeRF requires 3D supervision to be realized and does not contain semantic and instance information. In order to solve this problem, Nerflets [26] proposes that local radiation fields for efficient structure-aware 3D scene representation can be achieved with only 2D supervision, which is a new type of 3D scene representation. Nerflets represent a set of local neural radiance fields, collectively representing a scene, with each Nerflet possessing orientation, spatial position, and range. This allows for panoramic reconstruction within their respective ranges. In contrast to NeRF, Nerflets do not utilize a single large MLP to obtain the color and density of sampled points; instead, the influence weights of each Nerflet are combined. Specifically, for a sampled point, $N$ small MLPs generate $N$ densities and colors, one for each Nerflet, and then the final color and density of the sampled point are obtained through weighted averaging based on the influence weights. Nerflets can be utilized for 3D reconstruction, novel view synthesis, and direct scene editing. The indicator PSNR is also more than one point higher on the KITTI-360 and ScanNet datasets. The model framework is shown in Figure 4.

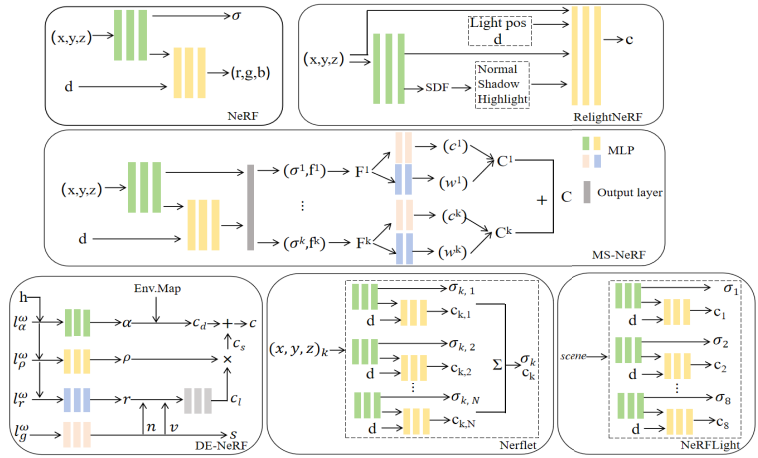

FIGURE 4. Comparison of core neural network architectures of NeRF, MS-RNeF, DE-NeRF, Nerflets, NeRFLight, and RelightingNeRF.

There are also NeRFLight [27] and RelightingNeRF [28] that change the underlying model architecture. The former extends the grid-based method by dividing the scene into eight different regions, effectively splitting NeRF into eight separate density fields and color fields. Each region corresponds to an MLP that outputs density and color. In the eight different regions, a shared feature grid is used, allowing features to be symmetrically placed at the center of the scene. This approach avoids the need to increase the number of features required for seamless reconstruction while reducing the number of feature values that need to be stored. The latter is proposed for handling unstructured photographs and introduces a new NeRF approach for free-viewpoint reconstruction. The model architecture is changed by inputting the sampling point position into the first MLP, which outputs SDF values and latent features. From the SDF values, we can obtain normals, shadow cues, and specular cues. By providing sampling point positions, lighting positions, and viewing directions, along with the latent features output by the first MLP, the second MLP produces the final color. This approach results in improved rendering quality. The framework of the above-mentioned paper that adjusts the NeRF model architecture is summarized in Figure 4. Comparison shows that researchers have derived more results and solved more problems based on the basic NeRF. Table 2 shows the indicator data of the above papers.

### C. USE PRE-TRAINED MODELS

Another extension of NeRF to improve image rendering quality is to use pre-trained models to optimize the NeRF model for better performance. Latent-NeRF [29] applies score distillation to a latent diffusion model, and then applies the entire process to the compact latent space of a pre-trained autoencoder. The specific implementation of Latent-NeRF is shown in Figure 5, where the MLP takes position and direction information as input and outputs the volume density and four pseudo color channels in a compact latent space.

TABLE 2. Three index values of related papers on adjusting the model structure.

| Model | PSNR | SSIM | LPIPS |
|---|---|---|---|
| MS-NeRF | 35.93 | 0.948 | 0.066 |
| DE-NeRF | 29.18 | 0.959 | 0.035 |
| Nerflets | 29.12 | - | - |
| NeRFLight | 31.41 | 0.968 | 0.039 |
| RelightingNeRF | 32.02 | 0.9727 | 0.0401 |

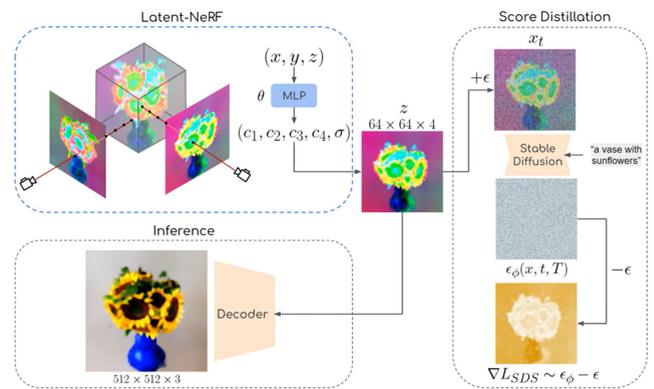

FIGURE 5. Latent NeRF trained by fractional distillation [29].

The output here differs from the basic NeRF in that the four pseudo color channels correspond to the four latent features used in the diffusion model. Then, a feature map $z$ is rendered from any viewpoint of the scene, and the resulting image is noised to generate a noisy image. The stable diffusion model is then used along with guidance from text and shape to denoise the image. Finally, the denoising process is backpropagated to the NeRF representation. Latent-NeRF is the first approach that allows coloring of the grid under the guidance of a pre-trained diffusion model, achieving impressive results.

NeRDi [30] is a single-view NeRF synthesis framework without 3D supervision that optimizes NeRF representation by minimizing the diffusion loss on arbitrary view rendering through a pre-trained image diffusion model. The specific model framework is depicted in Figure 6. Given an input image, and a pre-trained visual language model is used to introduce two language guides as the conditioned reflex input of the diffusion model. $s_0$ represents the image title containing the overall semantics, while $s^*$ is the text embedding extracted from the input image, which captures visual clues. These two language guides help improve the consistency of multi-view content. In addition, the paper also introduces a geometric loss based on estimated depth mapping to regularize the underlying 3D geometry of NeRF. These works can synthesize a more realistic new perspective image. The PSNR index on the DTU data set is twice as high as that of traditional NeRF, and it is also suitable for zero-shot NeRF synthesis.

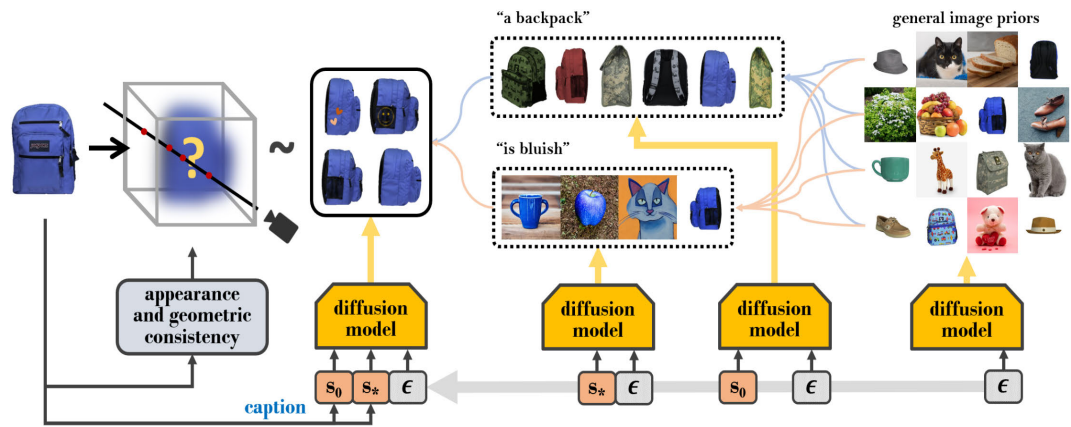

**FIGURE 6.** Overview of the NeRDi framework [30].

### D. HIGH RESOLUTION IMAGE PROCESSING

Although NeRF has achieved remarkable results, there is limited discussion among researchers regarding its limitations in high-resolution image processing. Because NeRF's MLP has a drawback: it struggles to capture high-frequency details, especially as the resolution increases. This leads to challenges such as an explosion of parameters and computational complexity, as well as misalignment of input data. Therefore, this section introduces the latest techniques for high-resolution processing in NeRF.

RefSR-NeRF [31] proposes a reference-guided super-resolution neural radiation field framework, which ultimately achieves high-quality new perspective synthesis by combining NeRF and RefSR-CNN models, and balances memory and speed. The framework process is shown in Figure 7. In the first stage, the sampling strategy of NeRF is modified. While NeRF uses batch ray sampling, RefSR-NeRF adopts patch ray sampling on the high-resolution input image to obtain a low-resolution image. This operation reduces memory storage requirements and improves the speed of NeRF. In the second stage, the obtained low-resolution image and the provided high-resolution guidance image are inputted to RefSR-CNN. This process yields feature maps from the detail-dominant information module and the degradation-dominant information module within RefSR-CNN. These two feature maps are then fused to reconstruct a high-resolution image, on the NeRF-LLFF data set, the PSNR index is nearly one point higher than the best paper.

**TABLE 3.** Pre-training and high-resolution related papers three index values.

| Model | PSNR | SSIM | LPIPS |
|---|---|---|---|
| NeRDi(single view) | 14.472 | 0.465 | 0.421 |
| RefSR-NeRF | 26.23 | 0.874 | 0.243 |
| AligNeRF | 25.51 | 0.734 | 0.306 |

AligNeRF [32] also discusses the limitations of training NeRF on high-resolution images through two improvements. 1) Combining MLP with convolutional layers to encode more neighborhood information and reduce the number of parameters; 2) Propose new strategies to address the impact of moving objects. The paper introduces a specific alignment-aware strategy that can correct the inaccuracies in alignment, which often lead to performance degradation. By improving the representational capacity of the neural radiance field, this strategy significantly enhances the quality of rendered images. AligNeRF not only improves rendering quality when training high-resolution images, but also does not significantly increase training time. From the out door data set, it is found that the PSNR index is better than the best mip-NeRF result, and is more than 3 points higher than NeRF. Table 3 shows the indicator data of the above papers.

### IV. SUMMARY AND OUTLOOK

Since its introduction in 2020, NeRF has provided a new perspective on view synthesis and 3D reconstruction, leading to rapid development and extensive discussions among researchers over the past three years. In the future, NeRF technology will continue to bring significant transformations to new domains.

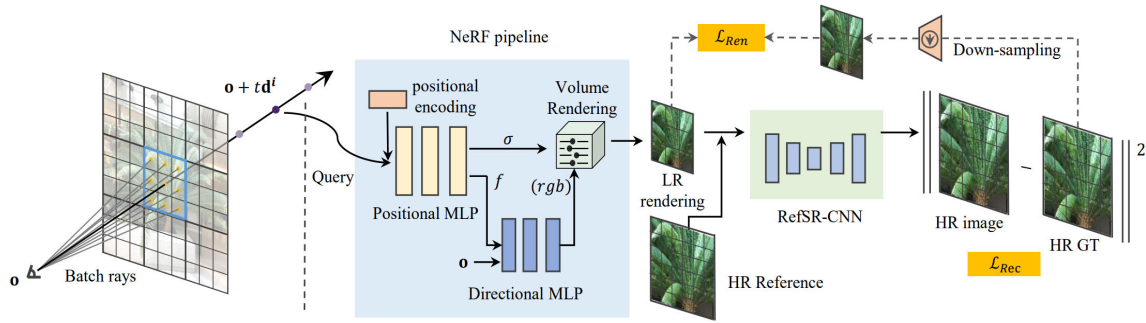

**FIGURE 7.** Overview of the RefSR-NeRF framework [31].

This article mainly discusses the image quality issues in the new perspectives of NeRF technology for view synthesis and 3D reconstruction. The factors that affect the rendering quality include insufficient training data, low image resolution, and high scene complexity. The article reviews the latest technologies from four aspects: prior guidance, modification of the base NeRF model structure, using pre-trained models, and high-resolution image processing. It introduces the principles and specific implementation methods of each technology for enhancing rendering quality based on NeRF.

NeRF technology is still rapidly evolving and there is a need for further advancements to enhance rendering quality in various real-world scenarios. Looking forward to future technological developments, we should focus on the following research directions.

### A. APPLIED TO LARGE-SCALE SCENES AND UNBOUNDED SCENES

Research on view synthesis and 3D reconstruction of large scenes and unbounded scenes has always attracted the attention of researchers. However, due to the complex configuration and dynamic changes of such scenes, some need to provide semantic logic or perform instance segmentation in advance, and others require information such as accurate depth estimation,which can limit the model's ability to express complex content. These types of scenarios could be addressed in the future by applying Normalized Device Coordinates (NDC) or Multi-Sphere Imagery (MSI). Additionally, an unknown pose NeRF model can be used to construct an end-to-end model for large-scale scene modeling.

### B. ACCELERATING RENDERING SPEED

In order to improve rendering quality, it is sometimes necessary to sacrifice rendering speed, such as increasing training time, adding more parameters, and enhancing image resolution. In the future, the goal is to improve rendering image quality without compromising rendering speed.

This can be achieved by first training the MLP model within NeRF using prior guidance to enhance rendering quality, and then distilling the knowledge to train a more compact MLP model. we can also Multi-threaded parallel processing scenarios; or use high-performance chips, because the development of chips and algorithms complement each other. By selecting chips that offer better matching in terms of performance, cost, power consumption, and other factors, the application of the algorithm can be further promoted.

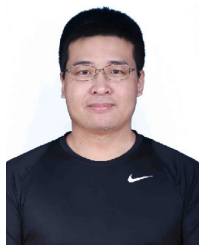

**SHUN FANG** (Senior Member, IEEE) received the Bachelor of Engineering degree from Northeast Electric Power University, in 2010, and the master's degree in engineering from Peking University, in 2016. He is currently the Vice President of Beijing Lumverse Technology Company Ltd., and the Director of the Lumverse Research Institute. He is also the Chief Scientist of future industry innovation tasks with the Ministry of Industry and Information Technology, China, the Chief Expert of technological innovation with Guangdong Provincial Department of Housing and Urban-Rural Development, and the Project Leader of the General Technology Platform Project, Zhongguancun National Independent Innovation Demonstration Zone. He is a Chartered Engineer with the Royal Institution of Engineers, U.K. Over the past decade, he has been actively involved as a Technical Expert in the development of Perfect World's ERA Engine and Lumverse's Lumverse 3D Engine. He has supported and serviced numerous large-scale 3D projects, generating over ten billion RMB in revenue. As a core member, he has participated in multiple research projects, including two national-level and four provincial-level projects, with a total funding scale exceeding 100 million RMB.

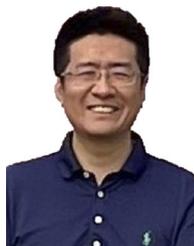

**MING CUI** received the bachelor's degree in engineering from Tsinghua University, in 1998, and the master's degree in engineering from Tsinghua University, in 2000. He was the Vice President of Technology with Perfect World, from 2004 to 2021, accumulating over 20 years of experience in game engine development. He is currently the Co-Founder and the CEO of Beijing Lumverse Technology Company Ltd. He is proficient in various game technologies and has led teams in the development of dozens of games. He has led teams in the independent development of large-scale server frameworks, 3D engines, and core business logic frameworks. The technical frameworks he designed and developed have contributed to over 50 billion RMB in revenue across multiple projects. He is a leading talent in the field of 3D engine development in China.

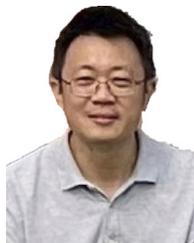

**XING FENG** received the bachelor's degree in engineering from Beijing University of Technology, in 2002, and the master's degree in engineering from the University of Abertay Dundee, U.K., in 2006. He studied under Dr. Henry, the Technical Lead of Sony PS2. He was the Senior Director of Perfect World, from 2012 to 2021. He is an expert of graphics rendering technology. He is currently the Co-Founder and the CTO of Beijing Lumverse Technology Company Ltd. With 15 years of experience in game engine development and management, he has participated in and led the development of multiple self-developed game engines. During his time with Perfect World, he led the ERA Team in servicing several heavyweight MMO games over a period of more than ten years, generating over ten billion RMB in revenue.

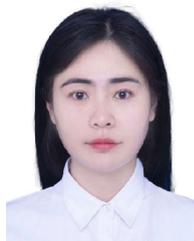

**YANNA LV** received the bachelor's degree in engineering from Shangqiu Normal University, in 2020, and the master's degree in engineering from Chongqing University of Technology, in 2023. She is currently a Researcher with the Lumverse Research Institute, focusing on research in 3D engines and artificial intelligence. She has participated in multiple municipal-level projects and was the Project Leader of one university-level project. She has published two Chinese core papers and received three university-level academic scholarships.

• • •